\documentclass{article} 

\usepackage{nips13submit_e,times}
\usepackage{url}
\usepackage{algorithm,algorithmic,amsmath,amsfonts,amssymb,mathrsfs,times}
\usepackage{graphicx}
\usepackage[font=small,labelfont=bf,tableposition=top]{caption}
\DeclareCaptionLabelFormat{andtable}{#1~#2  \&  \tablename~\thetable}

\def\transpose{{\mbox{\tiny T}}}

\def\mbi#1{\boldsymbol{#1}} 
\def\v#1{\mbi{#1}} 

\def\eqref#1{(\ref{#1})}
\def\secref#1{Sec.~\ref{#1}}
\def\reals{\mathbb{R}}

\newcommand{\laconic}{{Laconic}}

\title{Using Web Co-occurrence Statistics for Improving Image Categorization}

\author{
Samy Bengio, Jeff Dean, Dumitru Erhan, Eugene Ie \\
\textbf{Quoc Le, Andrew Rabinovich, Jonathon Shlens, and Yoram Singer} \\
\\
\texttt{\{bengio,jeff,dumitru,eugeneie\}@google.com}\\
\texttt{\{qvl,amrabino,shlens,singer\}@google.com}\\
\\
Google \\
Mountain View, CA, USA
}

\nipsfinalcopy 

\begin{document}

\maketitle

\begin{abstract}
Object recognition and localization are important tasks in computer vision.
The focus of this work is the incorporation of contextual information  in
order to improve object recognition and localization. For instance, it is
natural to expect not to see an elephant to appear in the middle of an
ocean.  We consider a simple approach to encapsulate such common sense
knowledge using co-occurrence statistics from web documents. By merely
counting the number of times nouns (such as elephants, sharks, oceans, etc.)
co-occur in web documents, we obtain a good estimate of expected
co-occurrences in visual data. We then cast the problem of combining textual
co-occurrence statistics with the predictions of image-based classifiers as
an optimization problem. The resulting optimization problem serves as a
surrogate for our inference procedure. Albeit the simplicity of the
resulting optimization problem, it is effective in improving both
recognition and localization accuracy. Concretely, we observe significant
improvements in recognition and localization rates for both ImageNet
Detection 2012 and Sun 2012 datasets.
\end{abstract}

\section{Introduction}
\label{section:introduction}
Object recognition from images is a challenging task at the intersection of
computer vision and machine learning. A major source of difficulty stems from
the fact that the number of object classes is large and it is easy to confuse
visually related objects. The bulk of the work on object recognition focused
on the task of identifying individual objects from a single or multiple
patches of the image. The existence of semantically related objects in a
single image is often sidestepped, though as the number of object classes
increases the potential of class confusion dramatically increases as well.

Few approaches were proposed in the computer vision literature to
incorporate contextual information, see for instance~\cite{galleguillos2008,
rabinovich2007, torralba2003}. Existing object classification methods that
incorporate context fall roughly into two categories. The first considers
global image features to be the source of context, thus trying to capture
class-specific features~\cite{torralba2003}. The second classifies objects
while taking into account the existence of the rest of the objects in the
scene~\cite{rabinovich2007}. The latter work used \emph{Google Sets} as a
source of information for co-occurrences of objects. Unfortunately this
source of information was inferior to using co-occurrence information
gathered from image training data. Research in the second setting has
further generalized in various ways. The work of Galleguillos et
al.~\cite{galleguillos2008} considers both semantic context, that is,
co-occurrences of objects as well as spatial relations between objects. In
\cite{chen2012}, the authors further extend the latter approach by proposing
an ``Object Relation Network'' that models behavioral relations between
objects in images. The aforementioned approaches typically require
laborious localization and labeling of objects in training images. To
mitigate this problem, the usage of unlabeled data through grouping of image
regions of the same context was proposed in~\cite{heitz2008}.

\begin{figure}[!ht]
\centering
\includegraphics[width=1\columnwidth]{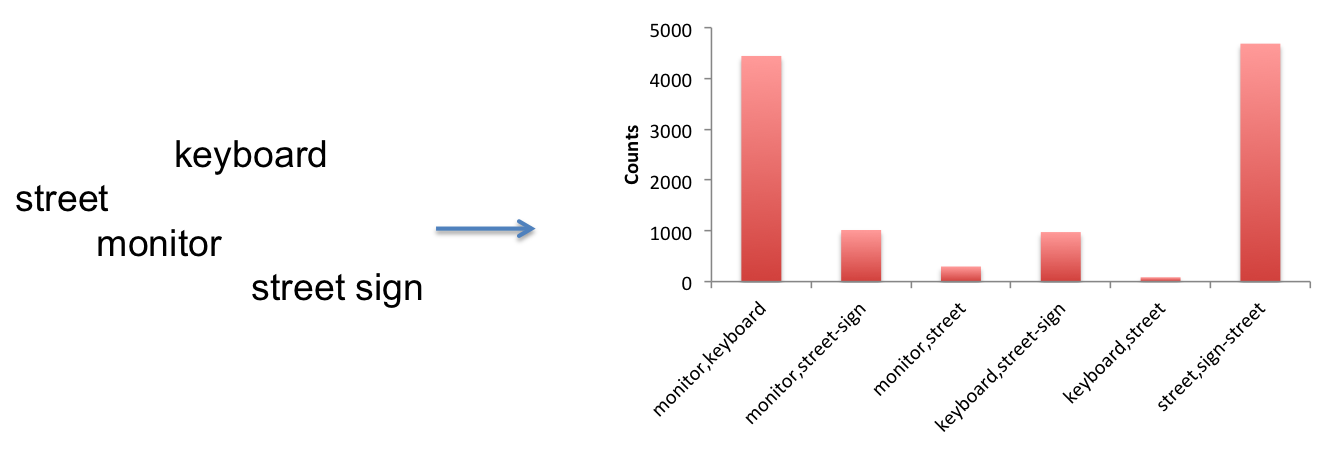}
\caption{\label{fig:intro} Results of co-occurrence counts for a set of
indoor and outdoor object pairs. The counts are computed from a small corpus
of text documents on the Web. Large counts correspond to object-pairs that
often co-occur together. It can be seen that objects found indoor also tend
to co-occur in text and similar behavior is observed for outdoor objects.
For example, {\bf monitor} and {\bf keyboard} have a high chance to appear
together in a text document while {\bf monitor} and {\bf street} do not.
This suggests that text co-occurrence is significantly correlated with visual
co-occurrence. Exploiting this fact is one of the contributions of this paper.}
\end{figure}

In this paper we present an approach that infuses an {\em external}
information source that captures the statistical tendency of visual objects
to co-occur in images. Rather than resorting to the image content itself for
the co-occurrence model, we use a large collection of textual data to count
the co-occurrences for any set of objects, as illustrated in
Figure~\ref{fig:intro}.

As mentioned above, most of prior work, including the cited papers, is built
upon the {\em visual} training set itself. To our knowledge, there was not a
substantial effort to incorporate external information from vastly different
information sources, such as web documents that is considered in this paper.
In contrast, modern speech recognition approaches have been using language
models for many years~\cite{rabiner}. In speech recognition systems, the
acoustic model is trained to recognize individual or short sequences of
phonemes. Its output is used by a ``decoder'' that incorporates language
statistics, typically in the form of a Markovian model whose parameters are
estimated from heterogeneous text sources. We propose to use an analogous
(yet simpler) approach for performing multiple object recognition.  While
images do not exhibit the rich morphological and syntactic structure of
natural languages, we show that the co-existence of semantically related
objects can still be leveraged.  Our approach indeed draws a parallel line
to speech recognition systems in which the acoustic model is combined with a
language model that is typically constructed from text-based sources.
However, due to the lack of temporal structure our decoding procedure
diverges from the dynamic-programming and cast the inference task as a
static optimization problem.

The end result is a general and scalable scheme that can be applied to
different sources of images while incorporating the statistical
co-occurrence model without the need of specialized training data.  We term
our approach \laconic\ as an acronym for {\em la}bel {\em con}sistency for
{\em i}mage {\em c}ategorization.

The rest of the paper is structured as follows. \secref{section:laconic}
describes the \laconic\ model and the underlying optimization that is used as
an inference tool. \secref{section:cooccurrences} describes the
co-occurrence model and how it is estimated from the web.
\secref{section:experiments} presents experimental results on two different
datasets (Sun12 and ImageNet) for which significant performance improvements
are shown when using \laconic. Finally, some conclusions and potential
extensions are provided in~\secref{section:conclusion}.

\section{The Laconic Setting}
\label{section:laconic}

\def\objective{{\cal Q}}
\def\extfield{{\cal E}}
\def\cooccur{{\cal C}}
\def\regularizer{{\cal R}}
\def\divergence{ {\cal D} }

We start by establishing the notation used throughout the paper. We denote
scalars by lower case letters, vectors by boldface lower case letters, e.g.
$\v{v}\in\reals^p$, and matrices by upper case letters. The transpose of a
matrix $A$ is denoted $A^\transpose$. Vectors are also viewed as $p\times 1$
matrices. Hence, the $2$-norm squared of a vector can be denoted as
$\v{v}^\transpose\v{v}$.  We denote the number of different object classes
by $p$.

The \laconic\ objective consists of three components: an image-based object
score which is obtained from an image classifier (see
\secref{section:experiments}), a co-occurrence score based on term proximity
in text-based web data (see \secref{section:cooccurrences}), and a
regularization term to prevent overfitting. We denote the vector of object
scores by $\v{\mu}\in\reals^p$ where the value of $\mu_j$ increases with the
likelihood that object $j$ appears in the image. We refer to this vector as
the {\em external field}. We denote the matrix of co-occurrence statistics
or object pairwise similarity by $S\in\reals_+^{p\times p}$. This matrix is
highly sparse and its entries are non-negative. Its construction is
described in \secref{section:cooccurrences}. We also optionally add domain
constraints on the set of admissible solutions in order to extract semantics
from the scores inferred for each label and as an additional mechanism to
guard against overfitting.

Abstractly, our inference amounts to finding a vector $\v{\alpha}\in\reals^p$
minimizing the following objective,
\begin{equation}
\label{eqn:laconic_objective}
\objective(\v{\alpha}|\v{\mu},S) =
  \extfield(\v{\alpha}|\v{\mu}) +
  \lambda \cooccur(\v{\alpha}|S) +
  \epsilon \regularizer(\v{\alpha})
  ~ ~ \mbox{s.t.} ~ ~ \v{\alpha}\in\Omega ~ ,
\end{equation}
where $\lambda$ and $\epsilon$ are hyper-parameters to be selected on a
separate validation set. For concreteness we describe the inference
procedure as a minimization task. The first term
$\extfield(\v{\alpha}|\v{\mu})$ measures the conformity of the inferred
vector $\v{\alpha}$ to the external field $\v{\mu}$. We tested the
following terms
\begin{eqnarray}
& -\v{\mu}^\transpose \v{\alpha} &
    \mbox{[linear]} \label{linearext:eqn} \\
& \sum_{j=1}^p \alpha_j \log\left(\frac{\alpha_j}{\mu_j}\right) +
      \mu_j - \alpha_j &
    \mbox{[relative entropy]} \label{relentext:eqn} ~ .
\end{eqnarray}
The linear score can be used in all settings, in particular, when the
outputs of the object recognizers take general values in $\reals^p$.
The relative entropy score can be used when the outputs are non-negative,
especially in the case where the outputs are normalized to reside in
the probability simplex. Such normalization often takes place by applying
a softmax function at the top layer of a neural network.

Shifting our focus to the second term $\cooccur(\v{\alpha}|S)$, we experimented
with the following scores,
\begin{eqnarray} &
  -\v{\alpha}^\transpose S \v{\alpha} &
  \mbox{[Ising]} \label{ising:eqn} \\ &
  \sum_{i,j} {S}_{i,j} \,\divergence(\alpha_i-\alpha_j) &
  \mbox{[difference]} \label{alphadiff:eqn} ~.
\end{eqnarray}
The Ising model assesses similarity between activation levels of object
category pairs. When $S_{i,j}$ is high, the two categories tend to co-occur
in natural texts underscoring the potential of similar co-occurrence in
natural images. Thus, the value of $\alpha_j$ will be ``pulled up'' by
$\alpha_i$ if the external field of the latter, $\mu_i$, is large. The sum
of \eqref{ising:eqn} and \eqref{linearext:eqn} is essentially the Ising
model~\cite{Ising}. In contrast to the physical model, we do not require an
integer solution. Alas, for general matrices $S$, the objective function
$\objective$ is not convex in $\v{\alpha}$. Thus, while we are able to use
classical optimization techniques, a convergent sequence is likely to end
in a local optimum. We revisit this issue in the sequel.

The divergence $\divergence : \reals^2 \rightarrow \reals_+$ assesses the
difference between the activation levels of two labels. Let $\delta$
denote the difference of an arbitrary pair $\alpha_i-\alpha_j$. We
evaluated and experimented with several options for $\divergence$,
based on $\ell_p$ norms and the Huber loss. We report result for the Huber
loss, defined as,
\begin{equation}
\label{huber:eqn}
\divergence_{\tiny \mbox{H}}(\delta) = \left\{
\begin{array}{ll}
\delta^2/2 & |\delta| \leq 1 \\
(|\delta|-1/2) & |\delta| > 1
\end{array}  \right. ~ .
\end{equation}
All the difference-based penalties we constructed are convex in
$\v{\alpha}$. However, difference-based penalties inhibit the activation
values of objects that tend to co-occur frequently in texts to become vastly
different from each other. This property while seemingly useful is in fact a
two-edged sword as it may create ``hallucination'' artifacts when solely one
of two highly-correlated labels appears in an image. Indeed, as our
experimental results indicate, by using multiple random starting points, the
Ising-like penalty yields better retrieval performance.

Finally, we need to describe the regularization component and the domain
constraints. Throughout our experiments we incorporated a $2$-norm
regularization, namely $\regularizer(\v{\alpha}) =
\v{\alpha}^\transpose\v{\alpha}$. In some of our experiments we cast an
additional requirement in order to find a {\em small} subset of the most
relevant labels. Concretely, we use a conjunction of a simplex constraint
and an $\infty$-norm constraint, yielding,
\begin{equation}
	\Omega = \left\{\v{\alpha}\,\mbox{ s.t. }\,\sum_j \alpha_j \leq N ~ , ~
  \|\v{\alpha}\|_\infty \leq 1 ~ , ~ \forall j: \alpha_j \geq 0\right\} ~ .
  \label{simplexinf:eqn}
\end{equation}
The above domain constraints relax the combinatorial requirement to have at
most $N$ object classes presented in the image. Since fractional solutions
are admissible, we often get more than $N$ non-zero indices in $\v{\alpha}$.
In order to find the optimum under the domain constraints of
\eqref{simplexinf:eqn} we need to perform gradient projection steps in which
each gradient step of \eqref{eqn:laconic_objective} is followed by a
projection onto $\Omega$. An efficient projection procedure is provided in
the supplementary material.

\section{Label Co-Occurrences}
\label{section:cooccurrences}

In order to estimate the prior probability of observing two labels $i$ and
$j$ in the same image, we harvested a sample of web documents, totalling a
few billion documents. For each document we examined every possible
sub-sequence (coined window) of consecutive words of length 20. We then
counted the number of times each label was observed along with the number of
co-occurrences of label-pairs within each window. We next constructed
estimates for the {\em point-wise} mutual information:
\begin{equation}
s_{i,j} = \log\left(\frac{p(i, j)}{p(i) p(j)}\right)\;,
\end{equation}
where $p(i,j)$ and $p(i)$ are the aforementioned counts normalized to
the probability simplex.

We discarded all pairs whose co-occurrence count was below a threshold.
Since web data is relatively noisy and our collection was fairly large,
even bizarre co-occurrences can be observed numerous times.
We thus set the threshold to $10^6$ appearances. We only kept pairs
whose point-wise mutual information was positive, corresponding to label-pairs
which tend to appear together. We then transformed the scores using the
logit function,
\begin{equation}
\label{eqn:logit}
S_{i,j} = \begin{cases}
\displaystyle\frac{1}{1 + \exp(- s_{i,j})}, & \mbox{if } s_{i,j} > 0 \\
0 & \mbox{otherwise.}
\end{cases}
\end{equation}

\begin{figure}[!ht]
\centering
  \centerline{\includegraphics[width=0.5\columnwidth]{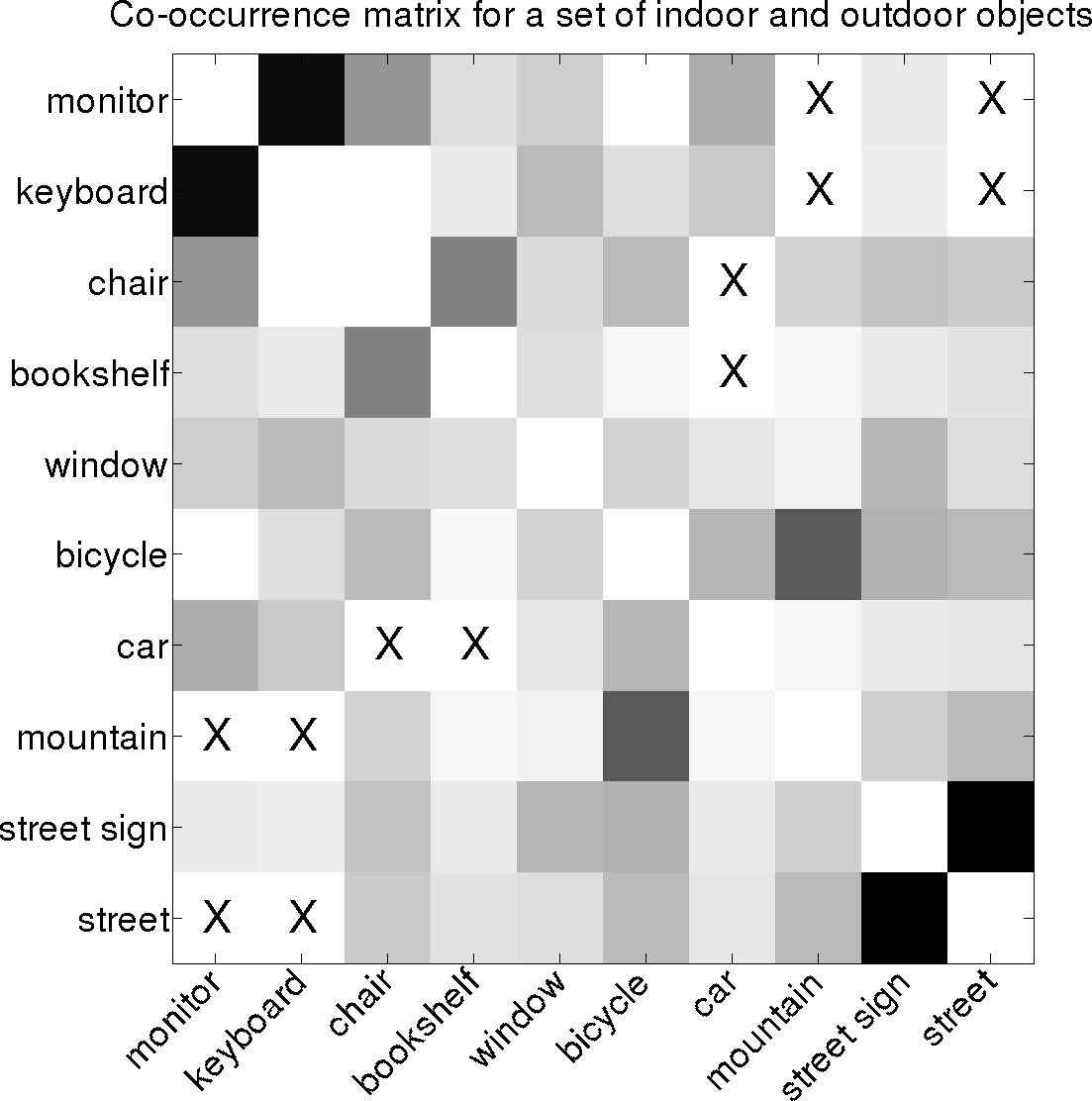}}
  \caption{\label{fig:10class_pmi} Graphical illustration of the
  co-occurrence scores for 10 object classes. Darker values show high
correlation; crosses mean negative correlation; the diagonal is not used.}
\end{figure}

In Fig.~\ref{fig:10class_pmi} we illustrate a typical matrix as processed
by equation~(\ref{eqn:logit}) for the following 10 classes: {\em chair}, {\em
bicycle}, {\em bookshelf}, {\em car}, {\em keyboard}, {\em monitor}, {\em
street}, {\em window}, {\em street sign}, {\em mountain}.  Dark squares
correspond to label-pairs of high mutual information, light squares to
low mutual information, and crossed squares to negative
mutual information, which are not used.
Pairs of objects that indeed tend to occur in the
same image, such as {\em keyboard} and {\em monitor}, attain higher scores
than object pairs that are not unlikely to be observed in the same image,
such as {\em street} and {\em keyboard}.

\section{Experiments}
\label{section:experiments}

In this section we report the results of two sets of experiments with public
datasets. The first dataset, Sun12, is small yet the largest of its kind as
each image is annotated with multiple labels. The second dataset, ImageNet
for detection, is much larger but most images are associated with a
single label.

In order to evaluate \laconic, we used the following metrics:
\begin{itemize}
\item Precision@$k$ is the number of correct labels returned in the top $k$
  positions, divided by $k$.
\item AveragePrecision is the mean of Precision@$k_i$ where $k_i$ is the
  position obtained by target label $i$, for all target labels.
\item Detection@$k$ is the number of correct detections returned in the top
  $k$ positions, where a detection is valid if both the label is correct and
  the returned bounding box overlaps at least 50\% with the target bounding
  box.
\end{itemize}
For each dataset, the training set was used to train a deep neural network
(DNN) for classification. The validation set was used to select all
hyper-parameters for the DNN and \laconic. The results for all the three
metrics above are reported for the test sets.

\subsection{Sun12}

Sun12 is a subset of the SUN dataset~\cite{xiao2010sun}, consisting of
fully annotated images from SUN. To our knowledge, Sun12 is the largest
publicly available dataset with multiple objects per image that spans thousands
of classes. There is a total of 16,856 images containing 165,271 objects from
3,765 classes. We split the dataset in our experiments into disjoint sets of
85\%, 10\%, and 5\% for training, validation and testing respectively. The
training set contains about 10 examples per object. The test set consists
of 798 images and is comprised of 7,761 annotated objects.

We created two types of co-occurrence matrices: one from web pages and
one from the Sun12 training set.  Since Sun12 contains multiple labels per
image, we built a co-occurrence matrix of object classes that appear
together in images from the training set. We report \laconic\ results using
the web-induced co-occurrence matrix, a matrix constructed from the Sun12
training set, and a convex combination of both. The mixture coefficient
was selected using the validation set.

For training the DNN we sampled around 1.5M cropped windows of varying sizes
from the training set, each of which contains at least 70\% of one target
object. These were then used to train a large scale convolutional network
similar to~\cite{KrizhevskySH12}, which won the last ImageNet
challenge. Concretely,
the model consisted of several layers, each of which performed 
a set of convolutions followed by local contrast normalization and
max-pooling. These layers were followed by several fully
connected linear and sigmoidal layers,
finally ending with a softmax layer with 3765 outputs, where each output
corresponds to one of the object classes in Sun12.
The full model was trained using stochastic gradient
descent combined with the AdaGrad algorithm~\cite{DuchiHS11} along with
the Dropout~\cite{dropout} regularization technique. We used
mini-batches of size 128 and an initial learning rate of 0.001.

At test time, we extracted the following seven patches at inference time
from the test image. The first is the largest possible {\em square} patch
centered at the middle of the image, which we refer to as ``the original
crop box''. The rest of the patches were the original crop box enlarged by
5\%, the original crop box shrunk by 5\%, and the original crop box
translated up, down, left and right by 5\%. The prediction scores of these 7 boxes
were then averaged.

We first ran a set of experiments in order to compare the various \laconic\
settings. Table~\ref{table:sun12:losses} compares several combinations of
the \laconic\ objective function described in \eqref{eqn:laconic_objective},
by varying the conformity term described in \eqref{linearext:eqn} and
\eqref{relentext:eqn}, or the co-occurrence term using either
\eqref{ising:eqn} or~\eqref{huber:eqn}. We also tested the incorporation of
domain constraints as described by \eqref{simplexinf:eqn}. The
combination consisting of a linear conformity term within the Ising model,
a $2$-norm regularization without further domain constraints seems to yield
good performance overall. These experiments used the co-occurrence matrix
constructed from web pages. Furthermore, since the Ising model for general
matrices is non-definite, we used 10 random initializations for $\v{\alpha}$
and selected the best solution according to \eqref{eqn:laconic_objective}.
Each inference of a single image took a few milliseconds on a standard Linux
machine. In comparison, naively using a conditional random field such as the
one described in~\cite{rabinovich2007}, would require the evaluation of about
${3765 \choose 5}$ combinations, which is greater than $10^{15}$. Instead,
only a few thousands evaluations were required when using \laconic.

\begin{table}[!ht]
\caption{\label{table:sun12:losses}Performance Results on the Sun12 Dataset
for various losses.}
\begin{center}
\begin{tabular}{|l|lll|}
\hline
Model vs Metric (\%)    & Precision@1 & Precision@5 & AveragePrecision \\ \hline
eqs~(\ref{linearext:eqn}) and~(\ref{ising:eqn}) & 86.2  & 53.1  & 76.3 \\
eqs~(\ref{relentext:eqn}) and~(\ref{ising:eqn}) & 86.3  & 52.8  & 76.2 \\
eqs~(\ref{linearext:eqn}),~(\ref{ising:eqn}) and~(\ref{simplexinf:eqn}) & 86.4  & 52.8  & 76.2 \\
eqs~(\ref{relentext:eqn}),~(\ref{ising:eqn}) and~(\ref{simplexinf:eqn}) & 85.8  & 52.6  & 75.9 \\
eqs~(\ref{linearext:eqn}),~(\ref{huber:eqn}) and~(\ref{simplexinf:eqn}) & 85.8  & 50.1  & 74.9 \\
\hline
\end{tabular}
\end{center}
\end{table}

Table~\ref{table:sun12} provides comparison results of the baseline model
(using the DNN only) to the various \laconic\ settings. In these
experiments we evaluated three settings for the co-occurrence matrix:
estimation using web data only, estimation using the Sun12 training set
only, and a convex combination of both matrices, termed Mixture-\laconic\ in
the table. As can be seen, the \laconic\ approach with the web
co-occurrence matrix gives significantly better performance than both the
baseline model and the \laconic\ approach with the Sun12 co-occurrence
matrix. The combination of both co-occurrence matrices gives slightly better
results overall.

\begin{figure}[!ht]
	\centerline{
		\includegraphics[width=1\columnwidth]{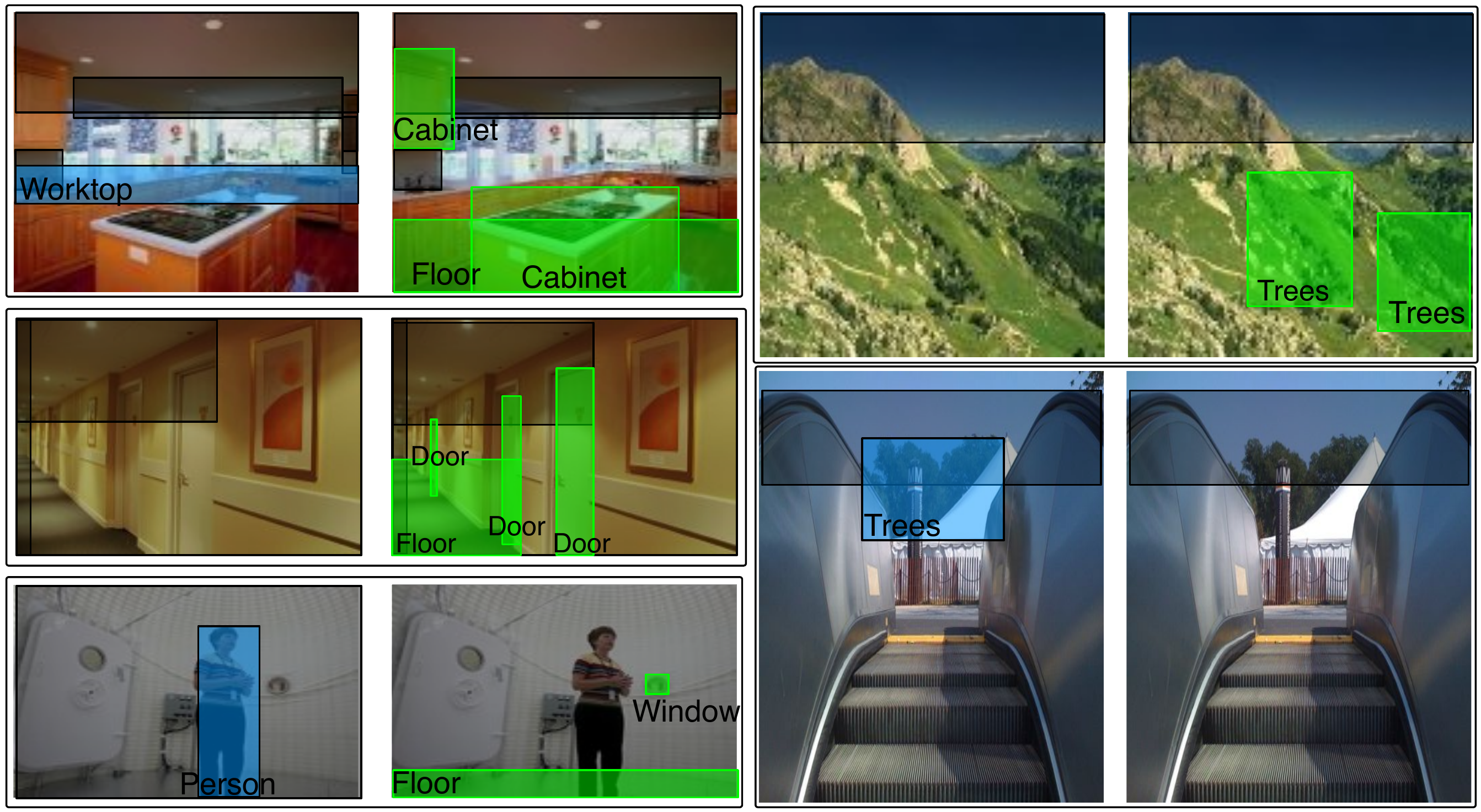}}
	\caption{\label{fig:sun12res_vis}
	Example classifications by the baseline system and \laconic\ with
	localizations. In each example pair, the baseline classifications are on
  the left, and 
	\laconic \ is on the right. Blue boxes with dark borders designate objects
	found only by the baseline. Green boxes with light borders designate
	objects found solely by \laconic. Black boxes designate objects
	found by both model.}
\end{figure}

\begin{figure}[!ht]
\centerline{\includegraphics[width=0.8\columnwidth]{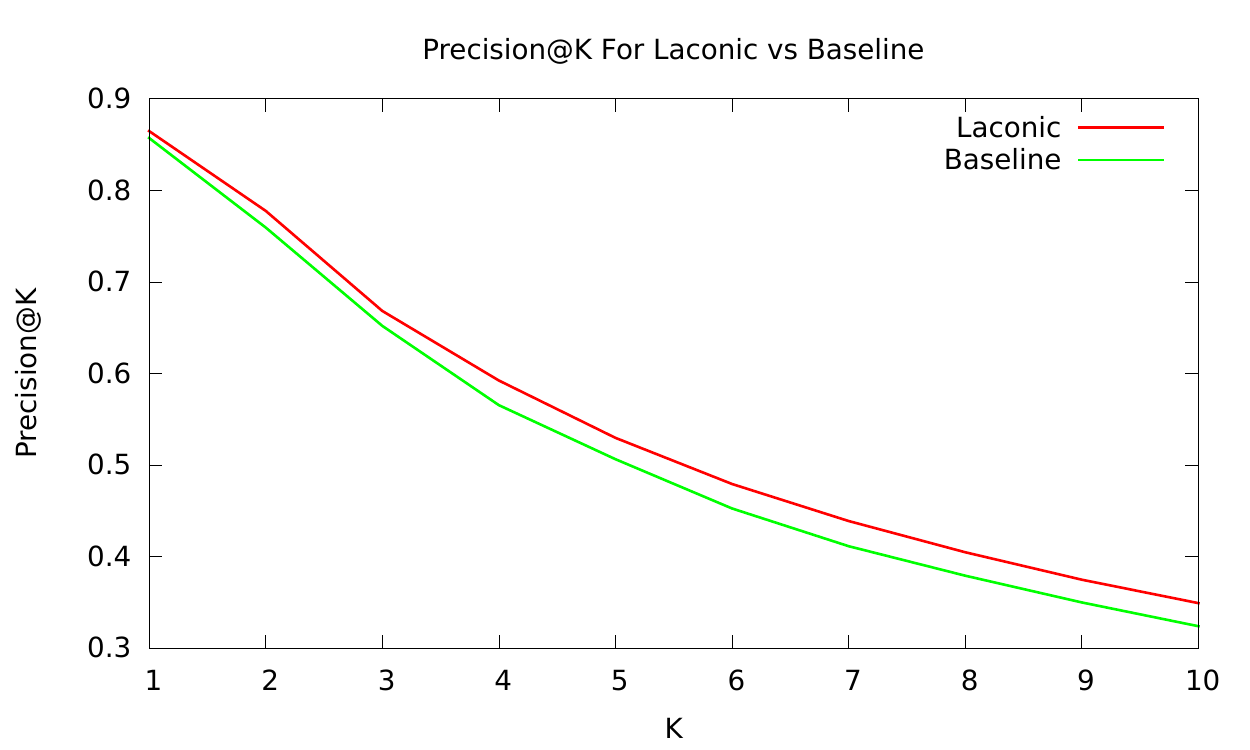}}
\caption{\label{fig:prec}Precision@K comparing \laconic\ to baseline on the
Sun12 dataset.}
\end{figure}

Examples of classification results by the baseline and \laconic\ are shown
in Figure~\ref{fig:sun12res_vis}, where we see that \laconic\ often surfaced
labels that were not found by the baseline, based on the additional
co-occurrence information. Figure~\ref{fig:prec} also compares \laconic\ to
the baseline for precision@$k$ for various values of $k$. Again, it shows
that \laconic's performance is better than the baseline for all values of
$k$.

\begin{table}[!ht]
\caption{\label{table:sun12}
Performance results on the Sun12 Dataset for different co-occurrence
sources.}
\begin{center}
\begin{tabular}{|l|lll|}
\hline
Model vs Metric (\%)    & Precision@1 & Precision@5 & AveragePrecision \\ \hline
Baseline & 85.7 & 50.6 & 75.0 \\
Web-\laconic\ & 86.2 & 53.1 & 76.3 \\
Train-\laconic\ & 85.5 & 50.8 & 75.1 \\
Mixture-\laconic\ & 86.1 & 53.7 & 76.6 \\
\hline
\end{tabular}
\end{center}
\end{table}

\subsection{ImageNet Detection Dataset}

Our second set of experiments is with the ImageNet dataset~\cite{deng2009}
(fall 2011 release). The entire dataset, typically used as a recognition
benchmark, has almost 22,000 categories and 15 million images. A subset of
this dataset is provided with bounding boxes which can be used for detection.
There are 3623 categories that have bounding boxes and the total
number of bounding boxes is 615,513. We divided the dataset into two
subsets of equal size: half for training and half for testing.

We trained a deep neural network on this dataset. The architecture of the
model is based on the model described in~\cite{le2012}.  It consists
of nine layers, has local receptive fields, employs pooling, local
contrast normalization, and a softmax output layer for multiclass
prediction.

To safeguard from overfitting, we added translated versions of the bounding
boxes and negative examples. First, for every instance of the bounding
boxes, we cropped 10 images, each translated up to 15 pixels in the
horizontal and vertical directions. These transformations yielded about
three million positive instances.  Next, we sampled negative patches and
treated them as a ``background'' category. To this end, we sampled random
patches from the training images and made sure that the overlapping area
with the correct bounding boxes is less than 30\% of the total area. We also
sampled examples that are difficult to classify by extracting patches that
overlap with the correct bounding boxes at the four corners. The total
number of negative (background) images is twelve million. Each of the above
image patches were resized to an image consisting of 100x100 pixels.
The total number of training images that were provide to learning algorithm
for the deep neural network was fifteen million.

Training was performed using stochastic gradient descent with mini-batches.
The resulting deep neural network was used for detecting objects in the
training set using a sliding window. The sliding window procedure was applied
at nine different scales where the $i$'th replica was of size $1.1^i$ of the
base window size.
For a specific scale, the overlap between one window to the next
window is 80\% of the size of the window. The windows were then resized
to images of 100x100 pixels each using cubic interpolation. The
patches were then fed to the deep network to obtain predictions. The
above scheme is naturally massively parallelized in the inference
step.




In order to asses the improvement of \laconic\ over the deep neural network
we performed evaluation of two tasks: detection and classification.  Since
the co-occurrence matrix is extracted from the web pages, it may not convey
any location information, and thus cannot help in localizing objects.  It is
therefore an interesting experiment on its own to see whether the external
text-based information can improve the detection accuracy.
Table~\ref{table:imagenet} compares the performance of the baseline network
with \laconic\ for two values of $k$: $k = 1$ and $k = 5$. The results show
that \laconic\ substantially improves both classification and detection. 

\begin{minipage}{\textwidth}
  \begin{minipage}[b]{0.49\textwidth}
    \centering
    \begin{tabular}{|l|ll|}\hline
      Metric (\%)      & Baseline & Web-\laconic\ \\ \hline
      Precision@1      & 24.8     & 34.8        \\
      Precision@5      & 8.9      & 10.9        \\ 
      AveragePrecision & 41.7     & 50.2        \\
      Detection@1      & 13.7     & 19.8        \\
      Detection@5      & 25.8     & 29.1        \\
      \hline
    \end{tabular}
    \vspace{0.5cm}
    \captionof{table}{\label{table:imagenet}Performance results on ImageNet}
  \end{minipage}
  \hfill
  \begin{minipage}[b]{0.49\textwidth}
    \centering
    \includegraphics[width=\textwidth]{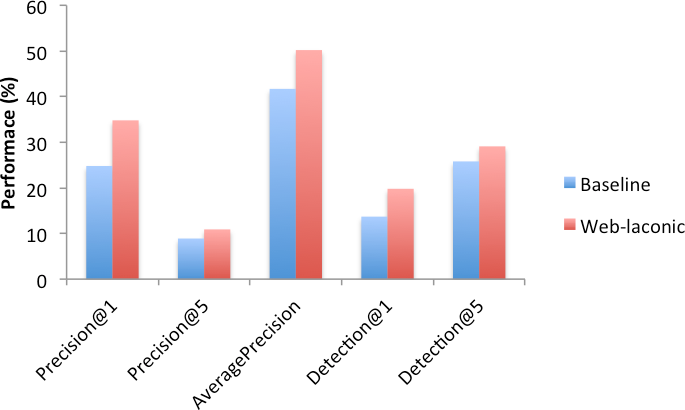}
  \end{minipage}
\end{minipage}

Note that even when there only a single object class is associated with
an image, \laconic\ can help in ``surfacing'' the correct class even if
it is not the most likely class according to the baseline DNN output.
Indeed, most images are typically under-labelled.
That is, there are more objects that appear in the images and thus
could have been added to the set of labels (object classes) of
the image. Since ImageNet was manually labeled where most images
were associated with a single label, the existence of additional
objects that tend to co-occur in natural text can improve the
classification accuracy. In other words, if the patch classifier assigned
high values to related object classes albeit not the single target
class, \laconic\ can increase the likelihood of the correct target
class, leveraging the co-occurrence information between the target
class and the aforementioned object classes.

Finally, it is worth noting that all results provided for the ImageNet
detection dataset, showing that Laconic was better than baseline, are
statistically significant with 99\% confidence, given the size of the
test set. On the other hand, the Sun12 dataset being so small, none of
the results are statistically significant, even at the 90\% level.

\section{Conclusion}
\label{section:conclusion}

Image classification becomes a difficult task as the number of object
classes increases. We proposed a new approach for incorporating side
information extracted from web documents with the output of a deep neural
network in order to improve classification and detection accuracy.  We
empirically evaluated our algorithm on two different datasets, Sun12 and
Imagenet, and obtained consistent improvements in classification and
detection accuracy on both datasets.

In future work we plan to go one step further and incorporate spatial
textual information. For instance, we could count the number of times we
observe sentences such ``the {\em chair} is {\em beside} the {\em table}''
or ``the {\em car} is {\em under} the {\bf bridge}'' and construct
triplets of the form $(object_1,relation,object_2)$, where the relation
expresses a spatial correspondence between the two objects. Such information
could potentially lead to a more comprehensive and accurate visual scene
analysis.

On the inference front, we plan to investigate alternative tractable
approaches. For instance, approaches based on belief propagation have
been used for related machine vision tasks. However, our preliminary
experiments with belief propagation yielded poor performance and are
thus not reported in the paper.


\bibliographystyle{plain}
\bibliography{biblio}

\begin{thebibliography}{10}

\bibitem{chen2012}
N.~Chen, Q.-Y. Zhou, and V.~K. Prasanna.
\newblock Understanding web images by object relation network.
\newblock In {\em Proceedings of the 21st World Wide Web Conference}, pages
  291--300, 2012.

\bibitem{deng2009}
J.~Deng, W.~Dong, R.~Socher, L.-J. Li, K.~Li, and L.~Fei-Fei.
\newblock Imagenet: A large-scale hierarchical image database.
\newblock In {\em {IEEE} Computer Vision and Pattern Recognition ({CVPR})},
  2009.

\bibitem{DuchiHS11}
J.~C. Duchi, E.~Hazan, and Y.~Singer.
\newblock Adaptive subgradient methods for online learning and stochastic
  optimization.
\newblock {\em Journal of Machine Learning Research}, 12:2121--2159, 2011.

\bibitem{galleguillos2008}
C.~Galleguillos, A.~Rabinovich, and S.~Belongie.
\newblock Object categorization using co-occurence, location and appearance.
\newblock In {\em {IEEE} Conference on Computer Vision and Pattern Recognition
  ({CVPR})}, 2008.

\bibitem{heitz2008}
G.~Heitz and D.~Koller.
\newblock Learning spatial context: Using stuff to find things.
\newblock In {\em {ECCV}}, pages 30--43. Springer, 2008.

\bibitem{dropout}
G.~E. Hinton, N.~Srivastava, A.~Krizhevsky, I.~Sutskever, and R.~R.
  Salakhutdinov.
\newblock Improving neural networks by preventing co-adaptation of feature
  detectors.
\newblock {\em arXiv preprint arXiv:1207.0580}, 2012.

\bibitem{Ising}
E.~Ising.
\newblock Beitrag zur theorie des ferromagnetismus.
\newblock {\em Z. Phys.}, 31:253–258, 1925.

\bibitem{KrizhevskySH12}
A.~Krizhevsky, I.~Sutskever, and G.~E. Hinton.
\newblock Imagenet classification with deep convolutional neural networks.
\newblock In P.~L. Bartlett, F.~C.~N. Pereira, C.~J.~C. Burges, L.~Bottou, and
  K.~Q. Weinberger, editors, {\em NIPS}, pages 1106--1114, 2012.

\bibitem{le2012}
Q.V. Le, M.A. Ranzato, R.~Monga, M.~Devin, K.~Chen, G.S. Corrado, J.~Dean, and
  A.Y. Ng.
\newblock Building high-level features using large scale unsupervised learning.
\newblock In {\em International Conference on Machine Learning}, 2012.

\bibitem{rabiner}
L.~Rabiner and B.-H. Juang.
\newblock {\em Fundamentals of Speech Recognition}.
\newblock Prentice Hall, 1993.

\bibitem{rabinovich2007}
A.~Rabinovich, A.~Vedaldi, C.~Galleguillos, E.~Wiewiora, and S.~Belongie.
\newblock Objects in context.
\newblock In {\em International Conference on Computer Vision ({ICCV})}, 2007.

\bibitem{torralba2003}
A.~Torralba.
\newblock Contextual priming for object detection.
\newblock {\em International Journal of Computer Vision}, 53(2):169--191, 2003.

\bibitem{xiao2010sun}
J.~Xiao, J.~Hays, K.~Ehinger, A.~Oliva, and A.~Torralba.
\newblock Sun database: Large-scale scene recognition from abbey to zoo.
\newblock In {\em 2010 {IEEE} Conference on Computer Vision and Pattern
  Recognition ({CVPR})}, pages 3485--3492, 2010.

\end{thebibliography}


\end{document}


\maketitle

\appendix
\section{An Efficient Projection Procedure} \label{appendix:projection}
In this appendix we describe an efficient projection procedure that
follows each gradient when we constrain the domain as described in
Section 2.
The procedure described here distills and
simplifies the procedure described in~\cite{PardalosKo90} and is
also provided for completeness. For brevity, we denote by $[p]$ the
set of integers $\{1,\dots,p\}$.

To simplify the derivation we denote the vector of label activations
after a gradient step by $\v{v}$. Naturally, $\v{v}$ does not adhere
with the domain constraints. Thus, we need to solve the following
projection problem,
\begin{equation} \label{genproj:eqn}
  \arg\min_{\v{\alpha}} \half \norm{\v{\alpha} - \v{v}}^2
  \; \mbox{ s.t. } \;
  0 \preceq \v{\alpha} \preceq 1 \, , \, \norm{\v{\alpha}}_1 \leq N \; .
\end{equation}
First, note that the ``box'' constraints ($0 \preceq \v{\alpha} \preceq 1$)
imply that if
\begin{equation} \label{vcap:eqn}
  \sum_j \max\{0, \min\{1, v_j\}\} \leq N
\end{equation}
then by bounding $\v{v}$ from above by $1$ and from below by $0$ we obtain the
optimal solution, namely, $\alpha_j = \max\{0, \min\{1, v_j\}\}$. If
\eqref{vcap:eqn} does not hold we need to use the algorithm described in the
sequel. Nonetheless, due to the positivity constraint we can set $\alpha_j=0$
for all indices $j$ for which $v_j \leq 0$ and drop the relevant components
when solving \eqref{genproj:eqn}.

We derive the algorithm for the general case by characterizing the form of
the solution using the Lagrangian of \eqref{genproj:eqn},
\begin{equation} \label{lagproj:eqn}
  {\cal L} = \half \norm{\v{\alpha} - \v{v}}^2 +
    \sum_j \lambda_j^1 (\alpha_j - 1) - \sum_j \lambda_j^0 \alpha_j +
    \theta \left(\sum_j \alpha_j - N\right) ~ ,
\end{equation}
where $\lambda_j^{0,1}$ and $\theta$ are non-negative Lagrange multipliers.
The KKT conditions for optimality imply that at the saddle point of
the Lagrangian the following holds for all indices $j$
$$
\frac{\partial {\cal L}}{\partial \alpha_j} =
  \alpha_j - v_j + \lambda_j^1 - \lambda_j^0 + \theta = 0 \;,
$$
or alternatively,
$$
  \alpha_j = v_j - \lambda_j^1 + \lambda_j^0 - \theta \;.
$$
Furthermore, at the optimal solution we need to satisfy the following
equalities for all indices $j$
\begin{equation} \label{optcond:eqn}
  \lambda_j^1 (\alpha_j - 1) = 0 \; \mbox{ and } \;
  \lambda_j^0 \alpha_j = 0 \; .
\end{equation}
Therefore, if at the optimal solution $\alpha_j$ is strictly smaller than $1$
then $\lambda_j^1$ must be zero. Similarly, if $\alpha_j>0$ then $\lambda_j^0$
must be zero. Thus, the optimal solution can be written in a closed form as
follows
\begin{equation} \label{optalpha:eqn}
  \alpha_j^\star =
    \max\left\{0, \min\left\{1, v_j - \theta^\star\right\}\right\} \; ,
\end{equation}
where $\theta^\star$ is the value obtained at the saddle point of the
Lagrangian~\eqref{lagproj:eqn}).
Let us now define the following function from $\reals_+$ to $\reals_+$,
\begin{equation} \label{dofalpha:eqn}
  \dofalpha(\theta) =
    \sum_j \max\left\{0, \min\left\{1, v_j - \theta\right\}\right\} \; .
\end{equation}

For clarity let us denote by $\v{\alpha}^\star$ the optimal value of
$\v{\alpha}$.  Since we checked that \eqref{vcap:eqn} does not hold, the
optimum is achieved at the boundary, $\norm{\v{\alpha^\star}}_1 = N$ and thus
$\dofalpha(\theta^\star) = N$. It then suffices to analyze the function
$\dofalpha$. First note that $\dofalpha$ is the sum of piece-wise linear
functions. Each summand is monotonically non-decreasing. Therefore,
$\dofalpha$ is piece-wise linear and monotonically non-decreasing.
Let us denote $\theta_{\max} = \max\{v_j\}$ and
$\theta_{\min} = \max\{0,\min\{v_j-1\}\}$.
For $\theta\in(\theta_{\min},\theta_{\max})$ the function $\dofalpha$ is
positive and monotonically decreasing.
Moreover, $\dofalpha(\theta_{\min}) > N$ and $D(\theta_{\max}) = 0$.
Therefore, there exists a unique value $\theta^\star$ for which
$D(\theta^\star) = N$.
call each point where $D$ changes its slope a {\em knot}.
For a given value of $\theta\in(\theta_{\min},\theta_{\max})$ let us denote
by $\alpha_j(\theta) = \max\{0,\min\{1,v_j-\theta\}\}$ and thus by definition
$\dofalpha(\theta) = \sum_j \alpha_j(\theta)$. We now partition the set of
indices of $[p]$ into three disjoint sets,
\begin{equation} \label{Is:eqn}
  \iz(\theta) = \{ j | \alpha_j(\theta) = 0\} \; , \;
  \im(\theta) = \{ j | 0 < \alpha_j(\theta) < 1\} \; , \;
  \io(\theta) = \{ j | \alpha_j(\theta) = 1\} \; . \;
\end{equation}
Alternatively, these sets can be written as
\begin{equation} \label{Is2:eqn}
  \iz(\theta) = \{ j | \theta \geq v_j\} \; , \;
  \im(\theta) = \{ j | v_j\!-\!1 < \theta < v_j \} \; , \;
  \io(\theta) = \{ j | \theta \leq v_j-1 \} \; .
\end{equation}
The partition into three sets facilitates a more explicit form
of $\dofalpha$,
\begin{equation} \label{dofalpha-exp:eqn}
  \dofalpha(\theta) = |\io| + \sum_{j\in \im} v_j - |\im|\theta ~ ,
\end{equation}
which is indeed a linear decreasing function at $\theta$.  Let us denote by
$\{\kappa_j\}$ the set of admissible knots in decreasing order, namely, a list
sorted in decreasing order of $\{v_j\} \cup \{v_j-1|v_j>1\}$. By definition
$D(\kappa_1) = 0$, $\iz(\kappa_1)=[p]$, $\im(\kappa_1)=\emptyset$, and
$\io(\kappa_1)=\emptyset$. We next describe an algorithm that brackets
$\theta^\star$, that is, two consecutive knots such that $\kappa_j <
\theta^\star \leq \kappa_{j+1}$.

For brevity we denote $\dofalpha_j \eqdef \dofalpha(\kappa_j)$. We need to
maintain an additional data structure which designates for each sorted knot
whether it corresponds to a value $v_i$ or to $v_i-1$. Note that the length of
the list is at most twice the dimension of $\v{v}$, namely, $2p$.  Let us
denote this list by $\v{b}$ where $b_j=+1$ if there exists $i$ such that
$\kappa_j=v_i-1$ and $b_j=-1$ otherwise. We have already characterized the
initial setting at $\kappa_1$. We therefore perform the following steps for
$j=2$ through the end of the sorted list. If $b_j=+1$ then we encounter a knot
where one of the components of $\v{v}$ moves from $\iz$ to $\im$. Note that
the slope of $\dofalpha(\theta)$ as described by \eqref{dofalpha-exp:eqn} is
the cardinality of $\im$. Therefore, the value of $\dofalpha$ at the newly
encountered knot can be derived from the previous knot as follows
\begin{equation} \label{iz2im:eqn}
\dofalpha(\kappa_{j}) \eqdef \dofalpha_{j} = \dofalpha_{j-1} +
  (\kappa_{j-1} - \kappa_{j}) |\im| ~ .
\end{equation}
In order to keep track of the value of $\dofalpha$ at the following knots we
also need to update the partition of the indices to the sets $\iz$, $\im$, and
$\io$. It suffices however to simply keep track of the {\em cardinality} of
$\im$. Let $\slope \eqdef |\im|$. Since $|\im\|$ increases by one when
$b_j=1$ we can simply update $\slope \leftarrow \slope + b_j$. We next
describe the case where $b_j=-1$ which corresponds to a transition from $\im$
to $\io$. First, note that the value at the newly encountered knot is computed
as before using \eqref{iz2im:eqn}. The sole difference is the update to the
slope since the cardinality of $\im$ decreased by one.  Nonetheless, we can
again perform the update $\slope\leftarrow \slope + b_j$.  Tracking the values
at the knots can stop once we encounter a value $\dofalpha_j \geq N$ which
implies that $\kappa_{j-1} < \theta \leq \kappa_j$.

It remains to show how to compute the value of $\theta^\star$ and obtain
$\v{\alpha^\star}$. Since $\theta^\star \in [\kappa_j,\kappa_{j-1})$ and
$\dofalpha(\theta^\star) = N$, we can simply perform a linear interpolation
to find $\theta^\star$
\[
\theta^\star = \kappa_j +
\frac{\dofalpha_j - N}{\dofalpha_j - \dofalpha_{j-1}}
  (\kappa_{j-1} - \kappa_j) ~ .
\]
Finally, to reconstruct $\v{\alpha^\star}$ we can simply use
\eqref{optalpha:eqn}. The pseudo-code of the algorithm after constructing
the $\v{\kappa}$ and $\v{b}$ through the calculation of $\theta^\star$ is
provided in Fig.~\ref{projalg:fig}.
\begin{algorithm}[t]
\caption{Finding the optimal offset $\theta^\star$} \label{projalg:fig}
\begin{algorithmic}[0]
  \REQUIRE $\v{\kappa}$ , $\v{b}$, $N$
  \STATE initialize: $\dofalpha_1 \leftarrow 0$, $\slope \leftarrow 0$
  \FOR{$j=2$ to $length(\v{\kappa})$}
  \STATE
  $ \dofalpha_j \leftarrow \dofalpha_{j-1} + (\kappa_{j-1} - \kappa_j) \slope $
  \IF{$\dofalpha_j \geq N$}
    \STATE break
  \ENDIF
  \STATE $\slope \leftarrow \slope + b_j$
  \ENDFOR
  \ENSURE
  $ \theta^\star = \kappa_j +
    \frac{\dofalpha_j - N}{\dofalpha_j - \dofalpha_{j-1}}
      (\kappa_{j-1} - \kappa_j) $
\end{algorithmic}
\end{algorithm}

{\small
\bibliographystyle{plain}
\bibliography{biblio}
}